\documentclass[letterpaper]{article} 
\usepackage{aaai25}  
\usepackage{times}  
\usepackage{helvet}  
\usepackage{courier}  
\usepackage[hyphens]{url}  
\usepackage{graphicx} 
\usepackage{natbib}  
\usepackage{caption} 
\usepackage{algorithm}
\usepackage{algorithmic}

\usepackage{newfloat}
\usepackage{listings}
\DeclareCaptionStyle{ruled}{labelfont=normalfont,labelsep=colon,strut=off} 
\floatstyle{ruled}
\newfloat{listing}{tb}{lst}{}
\floatname{listing}{Listing}
\usepackage{subcaption}
\usepackage{booktabs}
\usepackage{multirow}
\usepackage{color}
\usepackage{amssymb}
\usepackage{amsmath}
\usepackage{pifont}  

\title{Less is More: Token Context-aware Learning for Object Tracking}

\def\tracker{LMTrack}

\author{
    Chenlong Xu\textsuperscript{\rm 1},  Bineng Zhong\textsuperscript{\rm 1}\thanks{Corresponding author.},  Qihua Liang\textsuperscript{\rm 1}, Yaozong Zheng\textsuperscript{\rm 1} \\
    Guorong Li\textsuperscript{\rm 2},  Shuxiang Song\textsuperscript{\rm 1}
}
\affiliations{
    \textsuperscript{\rm 1}Key Laboratory of Education Blockchain and Intelligent Technology, Ministry of Education, \\
    Guangxi Normal University, Guilin 541004, China\\
    \textsuperscript{\rm 2}Key Laboratory of Big Data Mining and Knowledge Management, University of Chinese Academy of Sciences\\

    xuchenlong@stu.gxnu.edu.cn, bnzhong@gxnu.edu.cn,
    qhliang@gxnu.edu.cn, 
    yaozongzheng@stu.gxnu.edu.cn,\\
    liguorong@ucas.edu.cn,
    songshuxiang@mailbox.gxnu.edu.cn\\
}

\usepackage{bibentry}

\begin{document}

\maketitle

\begin{abstract}
Recently, several studies have shown that utilizing contextual information to perceive target states is crucial for object tracking. They typically capture context by incorporating multiple video frames. However, these naive frame-context methods fail to consider the importance of each patch within a reference frame, making them susceptible to noise and redundant tokens, which deteriorates tracking performance. To address this challenge, we propose a new token context-aware tracking pipeline named \textbf{\tracker}, designed to automatically learn high-quality reference tokens for efficient visual tracking. Embracing the principle of \textit{\textbf{L}ess is \textbf{M}ore}, the core idea of {\tracker} is to analyze the importance distribution of all reference tokens, where important tokens are collected, continually attended to, and updated. Specifically, a novel Token Context Memory module is designed to dynamically collect high-quality spatio-temporal information of a target in an autoregressive manner, eliminating redundant background tokens from the reference frames. Furthermore, an effective Unidirectional Token Attention mechanism is designed to establish dependencies between reference tokens and search frame, enabling robust cross-frame association and target localization. Extensive experiments demonstrate the superiority of our tracker, achieving state-of-the-art results on tracking benchmarks such as GOT-10K, TrackingNet, and LaSOT.
Code and models are available at \url{https://github.com/XuChenLong/LMTrack}.


\end{abstract}

\section{Introduction}

\begin{figure}[t]
\centering
\includegraphics[width=0.9\columnwidth]{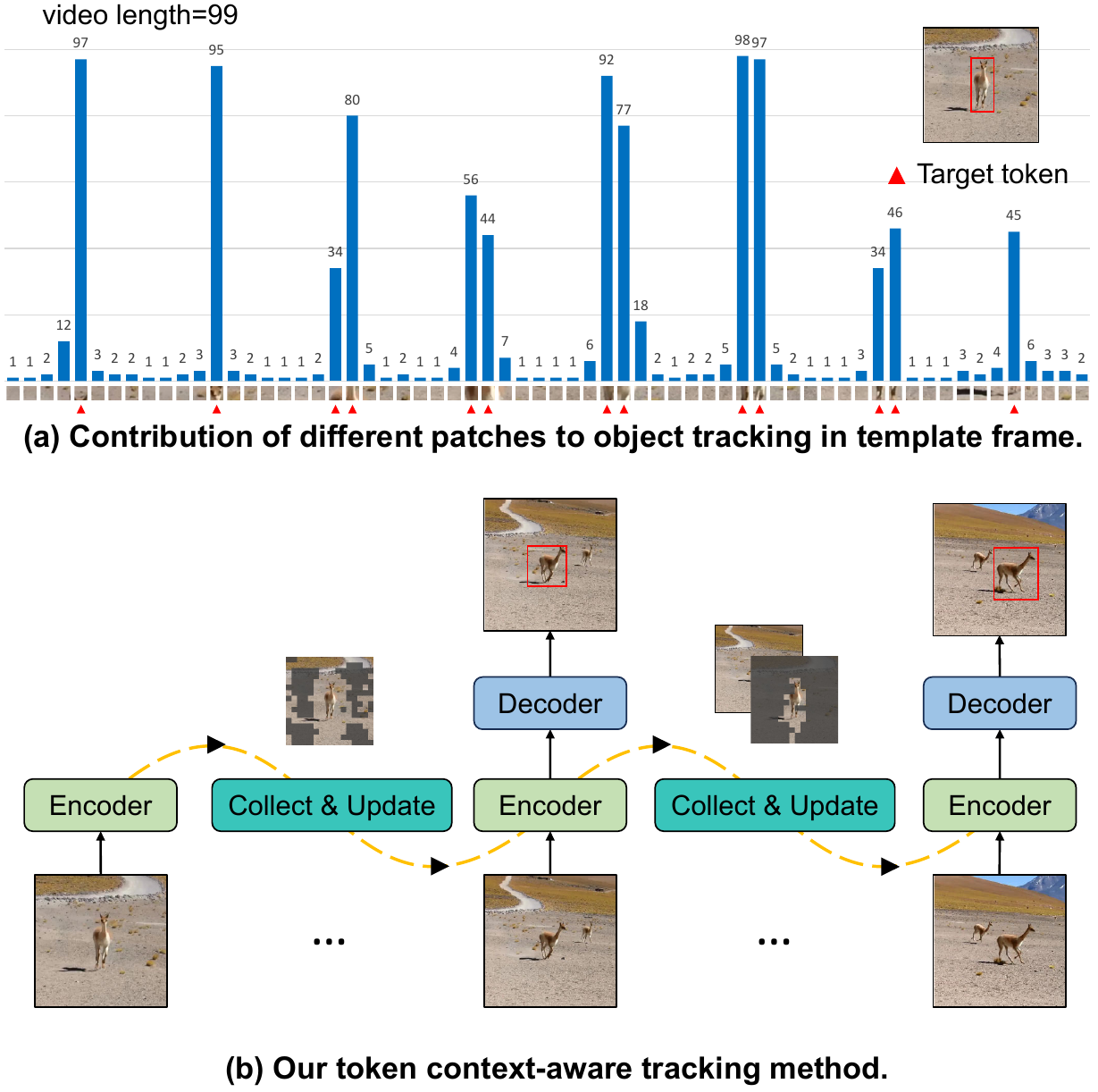} 
\caption{(a) Contribution (number of referenced) of different tokens to object tracking in a template frame. It can be observed that most background tokens are rarely referenced during the tracking process, while the target tokens retained as long-term reference cues. (b) Our token context-aware tracking method based on the token context memory module and the unidirectional token attention mechanism.
}
\label{index}
\end{figure}

Object tracking is a fundamental component of computer vision, designed to localize and track an arbitrary target within a video sequence based on its initial location.
To tackle this challenging task, recent research ~\cite{stark,STMTrack,mixformer,seqtrack,odtrack,Artrackv2} constructs high-performance tracking algorithms by exploring long-term contextual relationships. Typically, researchers achieve this within multiple video frames to capture contextual information. However, these naive methods have a significant drawback: they treat frame as the smallest units of context, neglecting that the importance of each patch in a reference frame is different for target localization in the search frame. This oversight makes them susceptible to redundant noisy information of the reference frame, thereby deteriorating tracking performance.

\textit{What type of reference cues play a dominant role in object tracking?} To answer this question, we design a simple Transformer tracker, consisting of a Transformer network (i.e., ViT~\cite{vit}), a classification head, and a regression head, to explore the impact of each patch within the template frame on object tracking (localization) throughout the entire video sequence. 
During inference, we first obtain the attention map from the search frame to the template frame and the classification score map. We then perform element-wise multiplication of the two maps and average along the template dimension to calculate the importance score of each token (patch) for target localization in the search frame. As shown in Fig.~\ref{index} (a), this gives us a distribution of importance scores for all reference tokens. We observe that most background tokens are rarely referenced during the tracking process and have minimal impact on the results, while target tokens are largely retained as long-term reference cues. This validates our motivation that a few high-quality tokens play a crucial role in the tracking process.

Based on the above findings, if we continue to allocate equal attention to all reference tokens during the entire tracking process, it will increase the model's perceptual and computational load, especially when dealing with complex scenarios. Adhering to the philosophy of \textit{less is more}, we design a simple yet effective token context-aware tracking pipeline named \textbf{\tracker}, which automatically learns high-quality reference tokens across timestamps for visual tracking.
As shown in Fig.~\ref{index} (b), the core idea of {\tracker} is to analyze the importance distribution of all reference tokens, where important tokens are collected, continually attended to, and updated.
Specifically, a novel Token Context Memory module is designed to dynamically collect and update high-quality spatio-temporal information of a target in an autoregressive manner, eliminating low-quality background tokens from the reference frames. This ensures that fewer reference tokens are used for accurate target localization in the search frames. Our approach discards the traditional frame-level context with redundant low-quality background information and instead uses a fine-grained token-level context to represent important reference cues across times-steps. This distinguishes our model significantly from other works~\cite{seqtrack,odtrack}. Furthermore, an effective Unidirectional Token Attention mechanism is designed to establish dependencies between reference tokens and the search frame in a unidirectional propagation manner, enabling robust cross-frame association and target localization.

Through this new modeling approach, we delegate the decision-making for target reference information to the tracker itself, rather than using handcrafted strategies for the tracker to passively accept reference frames. This empowers the tracker with an autonomous perception of reference cues, helping it adapt to target changes and preventing tracking drift.
The main contributions of this work are as follows:

    \begin{itemize}
    \item We propose a novel token context-aware tracking pipeline name {\tracker}. based on a Token Context Memory module. Unlike existing tracking methods with frame-level context, {\tracker} automatically collect and update high-quality token-context for visual tracking
    
    \item We introduce an effective unidirectional attention mechanism to establish dependencies between reference tokens and search frame in a unidirectional propagation manner, enabling robust cross-frame association and localization.
    
    \item Our approach achieves a new state-of-the-art tracking results on five visual tracking benchmarks, including LaSOT, TrackingNet, GOT10K, LaSOT$_{\rm{ext}}$, VOT2020.
    
    \end{itemize}

\section{Related Work}
\subsubsection{Traditional Tracking Framework.}
Visual object tracking has evolved significantly over the years, with traditional methods primarily relying on initial template approaches. Early methods~\cite{SiamFC,SiamFC++,SiamRPN++,Siamban} utilized Siamese networks to match the initial target template against candidate regions in subsequent frames. Although these methods effectively avoided tracker drift, they struggled to adapt to significant changes in the target's appearance. In recent years, the introduction of the transformer~\cite{attention} enables trackers~\cite{transt,ostrack} to enhance feature representation and matching capabilities, yet they continued to rely heavily on the initial template, limiting long-term tracking in ever-changing environments, which often requires addressing difficult target appearance issue. 
In contrast to these methods, we reformulate the object tracking as an important token collection task and aim to extend existing tracker to efficiently exploit the target temporal context.

\subsubsection{Temporal Context in Visual Tracking}
To handle the various appearance issues, many trackers have formulated the visual tracking issue as an online learning issue, in which the target appearance is adaptively updated using the temporal context of the previous frames. UpdateNet~\cite{updateNet} utilizes a custom network to fuse accumulated templates and generate a weighted updated template feature for visual tracking. ATOM~\cite {ATOM} adds IoU prediction branches to constrain template selection. STMTrack~\cite{STMTrack} updates dynamic templates at a fixed interval to counteract changes in the target appearance. STARK~\cite{stark} and Mixformer~\cite{mixformer} adopt an additional scoring head to verify whether the template contains the target, as the basis for selecting the template. SeqTrack~\cite{seqtrack} introduced a likelihood-based strategy that adopts the likelihood of generated tokens to select dynamic templates. RFGM~\cite{RFGM} selects the most appropriate template patches for the current search region, allowing for adaptation to variations. 

Nevertheless, the above tracking methods still suffer from the following limitations: 
(1) Most methods are designed to crop and update templates based on the bounding box. However, during online learning, they often incorporate a significant amount of noise or background, as the object typically does not occupy the entire bounding box. (2) Although they explore the temporal context to some extent, they update the template using manual approaches or additional discriminator models, failing to distinguish which contexts are essential for tracking.
To overcome these limitations, we propose {\tracker} based on the \textit{less-is-more} principle, which autonomously analyzes the importance distribution of all reference tokens, collecting and updating important target tokens as reference cues for subsequent video frames.

\begin{figure*}[t]
    \centering
    \includegraphics[width=0.85\linewidth]{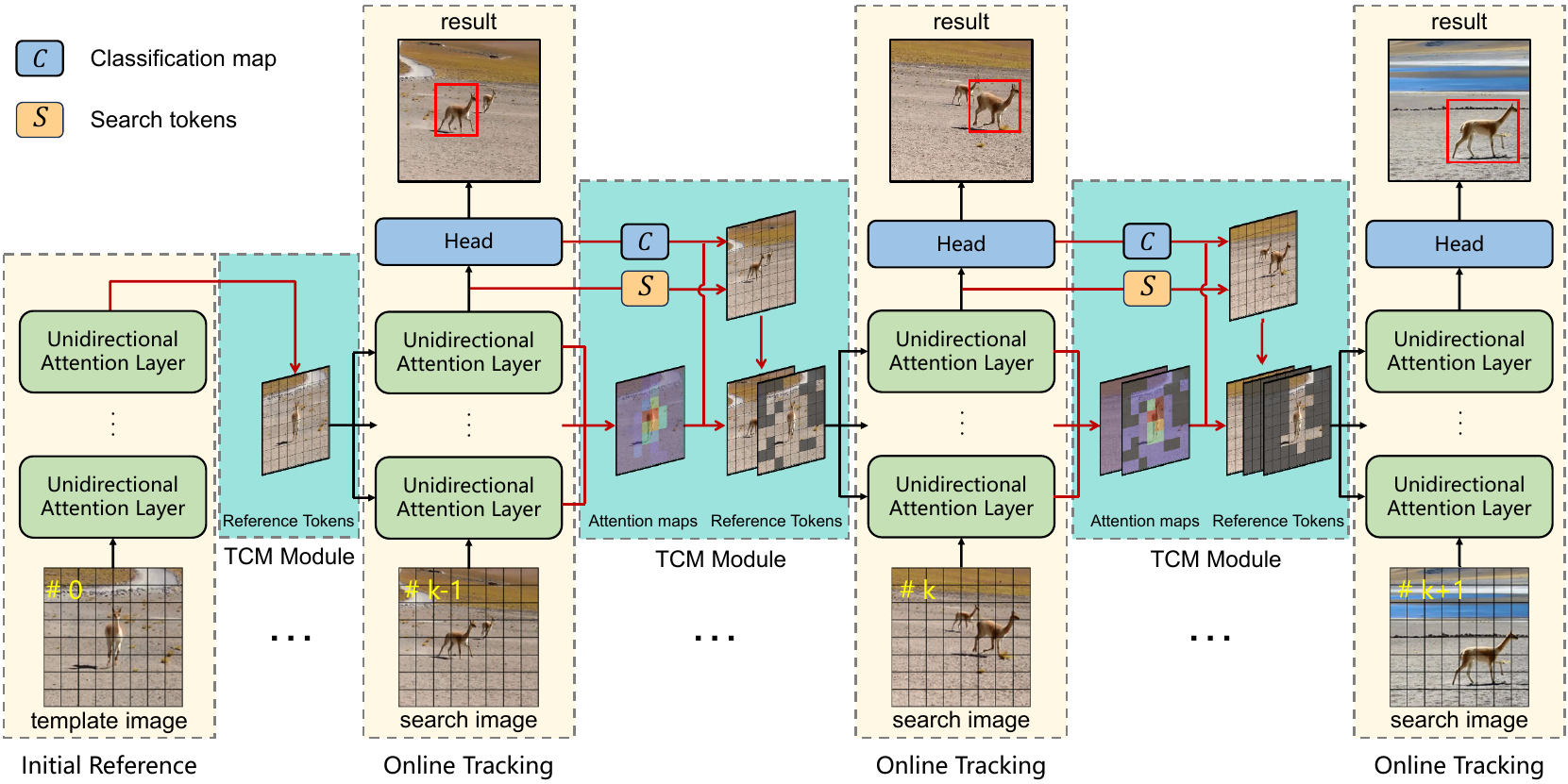}
        \caption{\textbf{The architecture of the \tracker}. {\tracker} consists of three parts, a backbone with unidirectional attention, a token context memory (TCM) module, and a prediction head. The input of tracking pipeline contains a video frame and reference tokens being collected. The TCM module utilizes classification maps and attention matrices to analyze the importance distribution of all reference tokens, then collect the important reference tokens according to this distribution. 
 }
    \label{fig:framework}
\end{figure*}

\section{Approach}

We propose a novel Token Context-Aware Tracking ({\tracker}) based on the \textit{less-is-more} principle. 
As depicted in Fig.~\ref{fig:framework}, {\tracker} comprises two key components: a novel Token Context Memory (TCM) module and an efficient unidirectional attention mechanism. 
This section begins with a brief introduction to our {\tracker} framework, then introduction of the proposed token context memory module and the unidirectional attention mechanism.

\subsection{Framework Formulation}
To provide a comprehensive understanding of our {\tracker}, it is essential to introduce our autoregressive token context-aware tracking framework.
Unlike previous approaches that only use the template image as a reference, our framework constructs the set of reference tokens based on all video frames for visual object tracking.
Therefore, we abandon the traditional approach of inputting image pairs consisting of template and search frames, opting instead for a single-frame input method. In this way, the corresponding reference tokens are adaptively collected and updated for each frame. In other words, we no longer strictly distinguish between template and search frames, instead, we treat each video frame equally, applying the same processes for target localization and reference token collection.
Specifically, we initialize the reference tokens $R_0$ using the template $I_0$, which shares the same backbone $g$ as the search frame $I_t (t>0)$. During the tracking process $f$ at time step $t$, the input consists of a search frame $I_t$ and the reference tokens $R_{t-1}$ being collected, while the output includes the predicted bounding box $B_t$ and the new reference tokens $R_t$.
The autoregressive tracking process $f$ is formalized as follows:
\begin{equation} \label{ar}
    \begin{split}
    &R_0 = f(I_0, \emptyset), t = 0,\\
    B_t, &R_t = f(I_t, R_{t-1}) = f(I_t, f(I_{t-1}, R_{t-2})), t > 0.\\
    \end{split}
\end{equation}


Specifically, {\tracker} utilizes the token context memory module to incrementally collect the relevant reference tokens from the initial frame to the current frame, which are then used as reference context for subsequent frames. During this process, {\tracker} discards irrelevant reference tokens, regardless of whether they originate from the initial or search frames. Our tracking framework consists of three components: a backbone network $g$ with a unidirectional attention mechanism, a prediction head $h$, and a reference context memory module. The representation process at time step $t$ is as follows:
\begin{equation} \label{eq:compose}
    \begin{split}
    S_t, A_t &= g(I_t, R_t), \\
     B_t, C_t &= h(S_t), \\
    R_t &= TCM(S_t, C_t, A_t, R_{t-1}).   \end{split}
\end{equation}
In this formulation, $S$ represents the search tokens processed by the backbone, while $A$ denotes the attention matrix that captures the distribution of importance between the reference tokens $R$ and the search tokens $S$. The backbone function $g$ employs the unidirectional attention mechanism rather than the traditional attention mechanism. 
The classification map $C$ corresponds to each search token in $S$. 

\subsection{Token Context-aware Tracking Pipeline}
In this section, we first introduce the Token Context Memory (TCM) module, which employs classification maps and attention maps to analyze the importance distribution of all reference tokens. Next, we present a unidirectional attention mechanism to effectively capture this importance distribution and enhance the efficiency of feature fusion. These components are designed to automatically learn high-quality reference tokens for visual tracking and consistently focus on these crucial tokens.

\subsubsection{Token Context Memory (TCM) Module}
\begin{figure}[t]
\centering
\includegraphics[width=0.9\columnwidth]{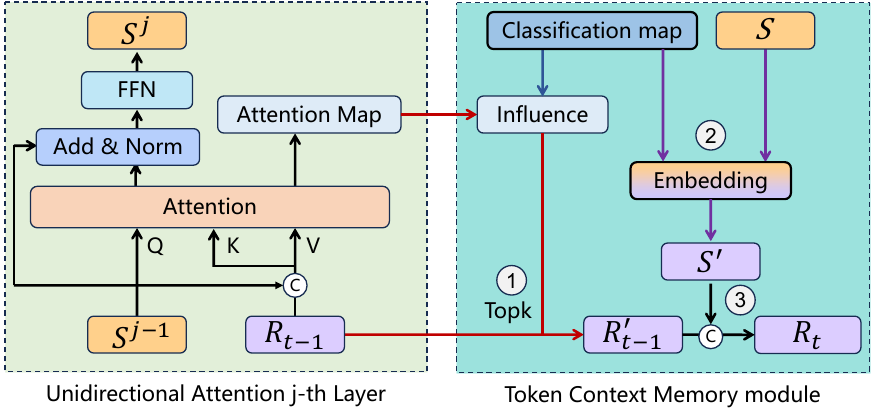}
\caption{
The unidirectional attention mechanism within the encoder layer is integrated with the token context memory module. The inputs to the unidirectional attention include search tokens and reference tokens. The token context memory module uses the attention map from unidirectional attention and predicted results to aggregate reference tokens.
}
\label{attention_layer}
\end{figure}

As illustrated in Fig.~\ref{attention_layer}, the token context memory module is divided into three steps: (1) Collect the important tokens from the existing reference tokens based on the classification map and the attention matrix; (2) Integrate the predicted classification map into the search tokens to be used as part of the reference tokens; (3) Update the reference tokens from steps (1) and (2) for subsequent tracking.

\textbf{STEP 1: Collect the important tokens from reference tokens.} Unlike trackers such as~\cite{seqtrack, mixformer}, which update template images based on temporal distance, {\tracker} gathers fine-grained, high-quality tokens from reference tokens that contain background and outdated target of redundant information according to their importance distribution. Fortunately, leveraging the powerful correlation calculations of the attention mechanism in the transformer architecture, {\tracker} can directly sample the cross-attention maps in each encoder layer, which primarily control the impact of the references within the encoder. This is combined with the tracking results to serve as a standard for collecting important reference tokens. Formally, we formulate the collection process as follows:
\begin{equation}
    \begin{split}
    &W = \sum^L_{j=1}A^j \times C, \\
    &R' = Topk(Rank(R, W)), \\
     \end{split}
\label{eq:k-score}
\end{equation}
where $A^j$ represents the cross-attention matrix between the reference tokens $R$ and search tokens $S^{j-1}$ in the $j$-th transformer layer, capturing the importance distribution of the current search tokens for reference tokens. The $C$ denotes the classification score map, reflecting the target distribution of the search tokens $S$. {\tracker} utilizes the $C$ and $A$ to assess the importance distribution of the reference tokens $R$. This process highlights the influence of each reference token $R$ on the target distribution in the search token $S$ and facilitates the differentiation of the importance of each reference token. Specifically, {\tracker} uses $A \times C$ as the refined metric for importance distribution, as opposed to merely summing the attention weights of all search tokens for each reference token. Subsequently, the importance distribution is aggregated across each encoder layer.
The $W$ signifies the importance distribution of each reference token $R$ relative to all search tokens $S$. {\tracker} retains the reference tokens $R$ corresponding to the $k$ largest $W$ as $R'$.
In the encoder layer, multiple importance distributions $W^m$ are generated due to multi-head attention, where $m=1, ..., M$ and $M$ is the number of attention heads. 
{\tracker} averages these importance distributions across all heads to obtain an overall importance score for the reference tokens $R$.

\textbf{STEP 2: Integrate the classification map and the search tokens.}
To enhance the representational power of the context memory, it is crucial to fully leverage the prediction results for the generated reference tokens.
Specifically, {\tracker} incorporates the category vector of the target, $E_{target}$, and the background, $E_{background}$. This integration is based on a binary classification score $C_{bin} \in [0, 1]$ obtained from $C$, $S$ is the output of the last encoder layer in the backbone $g$ and is used as a part of reference tokens $R$ in subsequent tracking.
This method enables the tracker to access not only potential reference tokens but also previous tracking results, thereby providing more comprehensive information than the original image alone.
The integration process is formally defined as follows:
\begin{equation}
    \begin{split}
    S' &= S + C_{bin}E_{target} + (1-C_{bin})E_{background}.
     \end{split}
\label{eq:generate}
\end{equation}

\textbf{STEP 3: Update the reference tokens.}
The {\tracker} obtains the reference tokens $R'_{t-1}$ identified in step (1) and the search features $S'_t$ that integrate the current tracking results from step (2). These components are then merged to form the reference tokens $R_t$ and update the $R_{t-1}$ for subsequent tracking. By continuously repeating these steps, the tracker constantly aggregates reference tokens that are valuable to the tracking process. Following this new modeling approach, {\tracker} adheres to the principle of \textbf{less-is-more}, achieving an autoregressive context-aware framework by selectively retaining fewer important reference tokens.

\subsubsection{Unidirectional Attention Mechanism}
The execution process of the TCM module shows that the accuracy of the importance distribution is crucial for effectively collecting the correct reference tokens. To enhance feature fusion efficiency and ensure the accuracy of the importance distribution, we employ a novel unidirectional attention mechanism to integrate reference features into search features, as illustrated in Fig.~\ref{attention_layer}. Compared to traditional attention fusion methods, this approach effectively prevents changes in the context representation of reference tokens caused by the influence of search tokens. For the backbone $g$ of a unidirectional attention encoder layer in Eq.~\ref{eq:compose}, the input consists of the $S$ (search tokens) from the previous encoder layer and the $R$ (reference tokens) updated in the previous tracking process. The operation is shown in the following formula:
\begin{equation} \label{eq:attention}
    \begin{split}
    S &= Softmax\left(\frac{Q_s[K_r; K_s]^T}{\sqrt{d_k}}\right)[V_r; V_s]\\
    &= Softmax\left(\frac{[Q_s{K_r}^T; Q_s{K_s}^T]}{\sqrt{d_k}}\right)[V_r; V_s],\\
    &= [A_r; A_s][V_r; V_s].
    \end{split}
\end{equation}
In the $j$-th unidirectional attention encoder layer, the previous output $S^{j-1}$ is projected into query, key, and value matrices $Q_s$, $K_s$, and $V_s$, respectively, while the reference tokens $R$ are projected into key and value matrices $K_r$ and $V_r$. Each unidirectional attention encoder layer uses the same $R$ within a single time step. The matrix $A_r$ represents the cross-attention matrix employed in the TCM module in Eq.~\ref{eq:k-score}. This unidirectional attention mechanism ensures that only the reference tokens $R$ affect the search tokens $S$, thereby maintaining a consistent representation of the reference tokens $R$ and reducing unnecessary computations. 

It is worth noting that we have implemented a context-aware token mechanism to automatically gather high-quality reference tokens using the TCM module. This process enhances the collection of crucial reference tokens for {\tracker}. During both training and inference, the TCM and the unidirectional attention mechanism work synergistically. The TCM improves reference token extraction within the unidirectional attention encoder by collecting appropriate tokens, while the unidirectional attention mechanism generates precise attention maps that benefit the effective collection of reference tokens.

\subsection{Head and Loss}
The output features $S$ from the encoder are input into a Fully Convolutional Network (FCN), which consists of $L$ stacked Conv-BN-ReLU layers for each output. The output of the FCN includes target classification score map \( \mathbb{R}^{\frac{H_x}{P} \times \frac{W_x}{P}} \), offset size \( \mathbb{R}^{2 \times {\frac{H_x}{P} \times \frac{W_x}{P}}} \) for compensating for discretization errors caused by reduced resolution, and normalized bounding box dimensions \( \mathbb{R}^{2 \times {\frac{H_x}{P} \times \frac{W_x}{P}}} \).

During training, both classification loss and regression loss are simultaneously employed. We utilize weighted focal loss~\cite{focalloss} for classification. For bounding box regression, we use predicted bounding boxes, $L_{1}$ loss, and generalized IoU loss~\cite{giou}. The total loss function is defined as:
\begin{equation} \label{eq2:loss}
    \begin{split}
    L = L_{cls} + \lambda_{iou}L_{iou} + \lambda_{L1}L_{1},
    \end{split}
\end{equation}
where \( \lambda_{iou} = 2 \) and \( \lambda_{L_1} = 5 \).

\begin{table}[t]
\centering
\caption{Comparison of model parameters, FLOPs, and inference speed.}
\resizebox{\linewidth}{!}{
\begin{tabular}{l|cccccc}
\toprule
Method & Type & Resolution & Params & FLOPs & Speed  & Device \\
\midrule
SeqTrack & ViT-B & $384\times384$ & 89M & 148G & 21$fps$ & 3090 \\
{\tracker} & ViT-B & $384\times384$ & 92M & 69G & 47$fps$ & 3090 \\
\bottomrule
\end{tabular} }
\label{tab:param}
\end{table}

\section{Experiments}
\subsection{Implementation Details}
\textbf{Training.}
We use ViT-base~\cite{vit} model as the visual encoder.
The training data includes LaSOT~\cite{lasot}, GOT-10k~\cite{got10k}, TrackingNet~\cite{trackingnet}, and COCO~\cite{coco}.
We employ the AdamW to optimize the network parameters with initial learning rate of $4 \times 10^{-5}$ for the backbone, $4 \times 10^{-4}$ for the rest, and set the weight decay to $10^{-4}$.
We set the training epochs to 300 epochs. $60,000$ search images are randomly sampled in each epoch.
The learning rate drops by a factor of 10 after 240 epochs.
The model is conducted on a server with two 80GB Tesla A100 GPUs, using a batch size of 16, where each batch consists of four search images and one template image.

\noindent
\textbf{Inference.}
In the initial stage of tracking, we use the first template to initiate the reference tokens. {\tracker} records the importance distribution for all reference tokens in each frame. It defaults to a reference update check every 400 frames and resets the importance distribution accordingly. When the sampled tokens from the search exceed the upper limit of the reference token length, we collect important reference tokens according to the importance distribution. The default maximum length of the reference tokens is twice the length of the search tokens. After collection, the length of the reference tokens is maintained at the initial length.

\subsection{Comparison with State-of-the-Art Trackers}
We demonstrate the effectiveness of {\tracker}, we compare them with state-of-the-art (SOTA) trackers on seven different benchmarks, including GOT-10K~\cite{got10k}, TrackingNet~\cite{trackingnet}, LaSOT~\cite{lasot}, LaSOT${\rm{ext}}$~\cite{lasot-ext}, VOT2020~\cite{vot2020}.

\begin{table*}[t]
    \centering
    \normalsize
    \resizebox{\textwidth}{!}{
    \begin{tabular}{l|ccc|ccc|ccc|ccc}
    \toprule
     \multicolumn{1}{c|}{\multirow{2}{*}{Method}} & \multicolumn{3}{c|}{GOT-10K$^*$} & \multicolumn{3}{c|}{LaSOT} & \multicolumn{3}{c|}{LaSOT$_{\rm{ext}}$} & \multicolumn{3}{c}{TrackingNet} \\
     \cline{2-13}
      & AO & SR${_{0.5}}$ & SR${_{0.75}}$ & AUC & P${_{\rm{Norm}}}$ & P & AUC & P${_{\rm{Norm}}}$ & P & AUC & P${_{\rm{Norm}}}$ & P \\
      \midrule
      SiamPRN++~\cite{SiamRPN++} & 51.7 & 61.6 & 32.5 & 49.6 & 56.9 & 49.1 & 34.0 & 41.6 & 39.6 & 73.3 & 80.0 & 69.4 \\
      DiMP~\cite{DiMP50} & 61.1 & 71.7 & 49.2 & 56.9 & 65.0 & 56.7 & 39.2 & 47.6 & 45.1 & 74.0 & 80.1 & 68.7 \\
      SiamRCNN~\cite{siamrcnn} & 64.9 & 72.8 & 59.7 & 64.8 & 72.2 & - & - & - & -  & 81.2 & 85.4 & 80.0\\
      Ocean~\cite{Ocean} & 61.1 & 72.1 & 47.3 & 56.0 & 65.1 & 56.6 & - & - & - & - & - & - \\
      STMTrack~\cite{STMTrack} & 64.2 & 73.7 & 57.5 & 60.6 & 69.3 & 63.3 & - & - & - & 80.3 & 85.1 & 76.7 \\
      TrDiMP~\cite{trdimp} & 67.1 & 77.7 & 58.3 & 63.9 & - & 61.4 & - & - & - & 78.4 & 83.3 & 73.1 \\
      TransT~\cite{transt} & 67.1 & 76.8 & 60.9 & 64.9 & 73.8 & 69.0 & - & - & - & 81.4 & 86.7 & 80.3 \\
      Stark~\cite{stark} & 68.8 & 78.1 & 64.1 & 67.1 & 77.0 & - & - & - & - & 82.0 & 86.9 & - \\
      KeepTrack~\cite{keeptrack} & - & - & - & 67.1 & 77.2 & 70.2 & 48.2 & - & - & - & - & - \\
      SBT-B~\cite{SBT} & 69.9 & 80.4 & 63.6 & 65.9 & - & 70.0 & - & - & - & - & - & - \\
      Mixformer~\cite{mixformer} & 70.7 & 80.0 & 67.8 & 69.2 & 78.7 & 74.7 & - & - & - & 83.1 & 88.1 & 81.6 \\
      TransInMo~\cite{TransInMo} & - & - & - & 65.7 & 76.0 & 70.7 & - & - & - & 81.7 & - & - \\
      OSTrack$_{384}$~\cite{ostrack} & 73.7 & 83.2 & 70.8 & 71.1 & 81.1 & 77.6 & 50.5 & 61.3 & 57.6 & 83.9 & 88.5 & 83.2 \\
      AiATrack~\cite{aiatrack} & 69.6 & 80.0 & 63.2 & 69.0 & 79.4 & 73.8 & 47.7 & 55.6 & 55.4 & 82.7 & 87.8 & 80.4 \\
      SeqTrack$_{384}$~\cite{seqtrack} & 74.5 & 84.3 & 71.4 & 71.5 & 81.1 & 77.8 & 50.5 & 61.6 & 57.5 & 83.9 & 88.8 & 83.6 \\
      GRM~\cite{GRM} & 73.4 & 82.9 & 70.4 & 69.9 & 79.3 & 75.8 & - & - & - & 84.0 & 88.7 & 83.3 \\
      VideoTrack~\cite{VideoTrack} & 72.9 & 81.9 & 69.8 & 70.2 & - & 76.4 & - & - & - & 83.8 & 88.7 & 83.1 \\
      ARTrack$_{384}$~\cite{ARTrack} & 75.5 & 84.3 & 74.3 & 72.6 & 81.7 & 79.1 & 51.9 & 62.0 & 58.5 & \underline{85.1} & 89.1 & 84.8 \\
      ODTrack$_{384}$~\cite{odtrack} & 77.0 & \underline{87.9} & 75.1 & \textbf{73.2} & \underline{83.2} & \underline{80.6} & 52.4 & 63.9 & 60.1 & \underline{85.1} & \textbf{90.1} & \underline{84.9} \\
      HIPTrack$_{384}$~\cite{HIPTrack} & 77.4 & 88.0 & 74.5 & 72.7 & 82.9 & 79.5 & \underline{53.0} &  \underline{64.3} & 60.6 & 84.5 & 89.1 & 83.8 \\
      AQATrack~\cite{AQATrack} & 76.0 & 85.2 & 74.9 & 72.7 & 82.9 & 80.2 & 52.7 & 64.2 & \underline{60.8} & 84.8 & 89.3 & 84.3 \\
      ARTrackV2~\cite{Artrackv2} & \underline{77.5} & 86.0 & \underline{75.5} & \underline{73.0} & 82.0 & 79.6 & 52.9 & 63.4 & 59.1 & \textbf{85.7} & 89.8 & \textbf{85.5} \\
      \midrule
      \textbf{{\tracker}$_{256}$} & 76.3& 87.1 & 73.9 & 69.8 & 79.2 & 76.3 & 49.0 & 59.6 & 55.8 & 84.2 & 89.0 & 82.8 \\
      \textbf{{\tracker}$_{384}$} & \textbf{80.1} & \textbf{91.5} & \textbf{79.0} & \textbf{73.2} & \textbf{83.4} & \textbf{81.0} & \textbf{53.6} & \textbf{64.7} & \textbf{61.5} & \textbf{85.7} & \underline{89.9} & 84.7 \\
    \bottomrule
    \end{tabular} }
    \caption{Comparison with state-of-the-arts on four popular benchmarks: GOT-10K \cite{got10k}, LaSOT \cite{lasot}, LaSOT$_{\rm{ext}}$ \cite{lasot-ext}, and TrackingNet \cite{trackingnet}. $^*$ denotes for trackers only trained on GOT-10K. The best two results are in \textbf{bold} and \underline{underline}, respectively.}
    \label{tab:results}
\end{table*}

\begin{table*}[t]
\centering
\resizebox{\textwidth}{!}{
\begin{tabular}{l|ccccccccccc|cc}
\toprule
& STM & SiamMask  & Ocean & D3S & AlphaRef & Ocean+ & STARK & SBT  & Mixformer &  SeqTrack & ODTrack$_{384}$ & \textbf{{\tracker}$_{256}$} & \textbf{{\tracker}$_{384}$}  \\
\midrule
EAO$(\uparrow)$ & 0.308 & 0.321 & 0.430 & 0.439 & 0.482 & 0.491 & 0.505 & 0.515 & 0.535 & 0.522 & \underline{0.581} & {0.550} & {\textbf{0.586}} \\
Accuracy$(\uparrow)$ & 0.751 & 0.624 & 0.693 & 0.699 & 0.754 & 0.685 & 0.759 & 0.752 & \underline{0.761} & - & \textbf{0.764} & {0.752} & {0.753} \\
Robustness$(\uparrow)$ & 0.574 & 0.648 & 0.754 & 0.769 & 0.777 & 0.842 & 0.819 & 0.825 & 0.854 & - & \underline{0.877} & {0.852} & {\textbf{0.895}} \\
\bottomrule
\end{tabular} }
\caption{State-of-the-art comparison on VOT2020~\cite{vot2020}. The best two results are in \textbf{bold} and \underline{underline}, respectively.}
\label{tab:expand_result}
\end{table*}

\textbf{GOT-10K.} GOT-10K~\cite{got10k} dataset is an extensive dataset comprising over 10,000 video segments, with 180 segments designated for testing. Following the official requirements, we only use the GOT-10k training set to train our model and evaluated the test results. As reported in Tab.~\ref{tab:results}, {\tracker} has achieved a remarkable state-of-the-art performance 80.1\% AO when compared to the previous best performance ARTrackV2 77.5\% AO. These results demonstrate that one benefit of our {\tracker} comes from token context-aware trackinng pipeline, which effect collects the token context during tracking.

\textbf{TrackingNet.} TrackingNet~\cite{trackingnet} is a large-scale dataset containing 511 videos and boasts a collection of over 30,000 videos with more than 14 million densely annotated bounding boxes. We evaluated {\tracker}$_384$ on its test set and achieved an impressive 85.7\% AUC on this large-scale benchmark.

\textbf{LaSOT.} LaSOT~\cite{lasot} dataset consists of 280 videos in its test set with an average length of 2448 frames. To assess the long-term tracking capabilities of {\tracker}. {\tracker}$_{384}$ surpasses the most of tracker, achieving a 73.2\% AUC. These results demonstrate that the TCM module can capture long-time contextual cues more efficiently.

\textbf{LaSOT$_{\rm{ext}}$.} LaSOT$_{\rm{ext}}$~\cite{lasot-ext} is an extended subset of LaSOT that includes 150 additional videos from 15 new categories. These new sequences introduce challenging tracking scenarios, such as occlusions and fast-moving small objects. {\tracker} gets a 53.6\% AUC, 64.7\% P$_{Norm}$ and 61.5\% P, outperforming the ARTrackV2 by 0.7\%, 1.3\%, 2.4\%, respectively. This  demonstrates the robustness of {\tracker} in handling these difficult scenarios.

\textbf{VOT2020}. VOT2020~\cite{vot2020} contains 60 challenging sequences and uses binary segmentation masks as the groundtruth. We use Alpha-Refine~\cite{Alpha-Refine} as a post-processing network to predict segmentation masks. As shown in Tab.~\ref{tab:expand_result}, {\tracker}$_{256}$ and {\tracker}$_{384}$ achieve the EAO results of 55\% and 58.6\% on mask evaluations, respectively, demonstrating the effectiveness of the token context-aware approach.


\subsection{Ablation and Analysis.}
In this section, we perform a detailed analysis of the key components of {\tracker}$_{256}$. In all our experimental studies, we adhere to the GOT-10K test protocol.

\begin{table}[h]
\centering
\small
\resizebox{\linewidth}{!}{
\begin{tabular}{l|c|c|c|ccc}
\toprule
\# & Attention & autoregressive & Update & AO(\%)\\
\midrule
1 & bidirectional & $\times$ & - & 73.0\\
2 & unidirectional & $\times$ & - & 73.9 \\
3 & unidirectional & $\times$ & update template & 74.1\\
4 & unidirectional & $\times$ & TCM & 75.0\\
5 & unidirectional & \checkmark & update template & 75.6\\
6 & unidirectional & \checkmark & TCM & 76.3\\
\bottomrule
\end{tabular}
}
\caption{Ablation experiment about {\tracker} in GOT-10K.}
\label{tab:ablation}
\end{table}

\textbf{The Unidirectional Attention.} 
To evaluate the impact of the unidirectional attention mechanism described in Eq.~\ref{eq:attention}, we conducted experiments comparing different attention mechanisms, as shown in Tab.~\ref{tab:ablation}. 
The bidirectional attention method processes both the search and template images simultaneously, whereas the unidirectional attention method only takes the search image and reference tokens from the initial template image as inputs. Observations from the first and second rows indicate that the unidirectional attention mechanism prevents noise from propagating from the search to the reference, resulting in a 0.9\% increase in average overlap (AO). Additionally, unidirectional attention improves fusion efficiency. As seen in Tab.~\ref{tab:param}, unidirectional attention significantly enhances inference speed when using the same template/reference token sizes.
This shows that unidirectional attention not only prevents noise propagation but also eliminates duplicate modeling of template features.

\textbf{Autoregressive Tracking.} 
We compare the different tracking method on the performance. Traditional methods require cropping the template based on previous results and extracting features with the backbone at each update. This approach uses only $R_0 = f(I_0, \emptyset), t=0$ and does not employ $B_t, R_t = f(I_t, R_{t-1}), t>0$. As shown in Tab.~\ref{tab:ablation} (rows three to six), autoregressive feature extraction outperforms traditional methods, achieving a 1.5\% and 1.3\% improvement in AO over the template update method and the TCM module, respectively. Traditional methods lack reference tokens for accurately modeling target information in subsequent templates, limiting their ability to effectively disseminate target information throughout the video. In contrast, the autoregressive method leverages reference tokens updated in previous tracking steps, adapting more effectively to subsequent tracking processes and eliminating the need for redundant operations on new template images.

\textbf{Token Context Memory Module.} 
The Token Context Memory (TCM) module is designed to enhance reference token representation by adhering to a \textit{less-is-more} principle, selectively retaining a less number of important reference tokens. This selective retention strategy allows for more precise updates of the reference tokens and updates each reference token independently, instead of modifying the entire template image. To evaluate the impact of various update strategies, we adopt {\tracker} as our default configuration. The results are shown in Tab.~\ref{tab:ablation} (rows three to six), demonstrating that employing more fine-grained operations achieves a 0.7\% and 0.9\% improvement in AO with and without autoregressive tracking, respectively. This shows that {\tracker} is able to constantly aggregate reference tokens that are valuable for the tracking process.


\begin{figure}[t]
    \centering
    \includegraphics[width=0.85\linewidth]{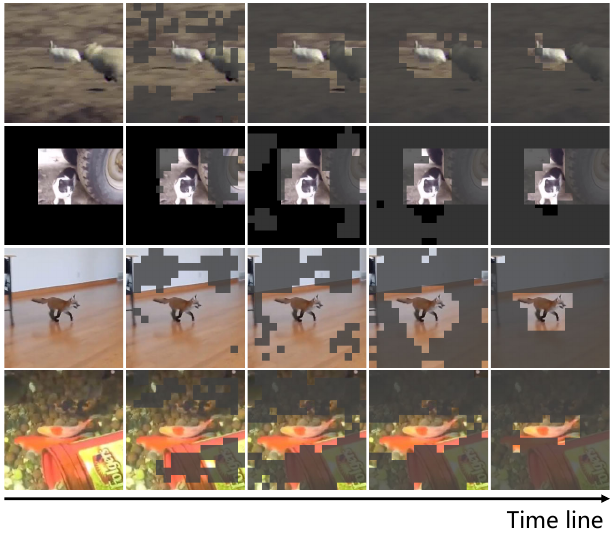}
        \caption{The TCM module visualization shows the collection of significant reference tokens over time. The highlighted part indicates retained tokens for ongoing tracking, illustrating the module's ability to filter out invalid tokens.}
    \label{fig:ce}
\end{figure}

\subsection{Visualization and Qualitative Analysis}
\textbf{Visualization of the TCM Module:} Fig.~\ref{fig:ce} depicts the process by which our TCM module extracts significant reference tokens from the same frame over time. As time progresses, most background tokens become less significant, with fewer reference tokens primarily describing the target's appearance. Our autoregressive tracking method effectively captures the reference tokens associated with the target, even in challenging circumstances that involve variations in appearance and potential distractions.This illustrates a notable adaptability in accurately identifying the reference tokens of the tracking target.


\begin{figure}[t]
    \centering
    \includegraphics[width=0.85\linewidth]{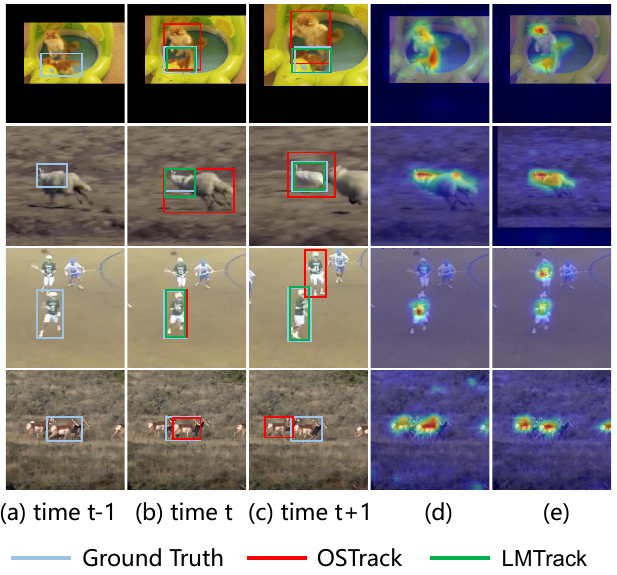}
        \caption{Comparison of the response between \tracker and OSTrack. (a)-(c): Search regions with predict boxes. (d): Attention maps of {\tracker}. (e): Attention maps of OSTrack.}
    \label{fig:ce}
\end{figure}

\textbf{Compare Response Between reference tokens and Template:} When faced with these difficult scenarios, the template image may be unreliable. To address these problems, our method uses reference tokens instead of the original template image to guide object tracking. In Fig.~\ref{fig:ce}, we compare the attention maps generated by the reference tokens of {\tracker} and the template of OSTrack~\cite{ostrack}. Unlike OSTrack, which uses the template image for guidance, {\tracker} employs historical search features to generate reference tokens and collects important reference tokens via the TCM module, effectively resisting various appearances and distractions. 
By observing tracking results frame-by-frame and attention response maps, we find that the attention of OSTrack becomes distracted when similar objects are present in the search image, leading to erroneous tracking. In contrast, {\tracker} can maintain focus on the target by using reference tokens that have proven important in previous tracking steps to keep attention on the target. 


\section{Conclusion}
This study introduces the Token Context-Aware Tracker ({\tracker}), which is based on the principle that less tokens with tracker attention play a more important role in the results. {\tracker} comprises two key components: The Token Context Memory (TCM) module and the unidirectional attention mechanism in the encoder layer. The TCM module analyzes the importance distribution of all reference tokens, collecting, attending to, and updating the important ones. Additionally, {\tracker} adopts the unidirectional attention mechanism to establish dependencies between reference tokens and search frame in a unidirectional propagation manner. Thus, our approach discards the traditional frame-level context and achieves a fine-grained token-level context to represent important reference cues across time steps. 

\section{Acknowledgements}
This work is supported by the National Natural Science Foundation of China (No.U23A20383 and 62472109), the Project of Guangxi Science and Technology (No.2024GXNSFGA010001 and 2022GXNSFDA035079), the Guangxi ”Young Bagui Scholar”Teams for Innovation and Research Project, the Research Project of Guangxi Normal University (No.2024DF001), and the Innovation Project of Guangxi Graduate Education (YCSW2024147).

\bibliography{aaai25}

\begin{thebibliography}{39}
\providecommand{\natexlab}[1]{#1}

\bibitem[{Bai et~al.(2024)Bai, Zhao, Gong, and Wei}]{Artrackv2}
Bai, Y.; Zhao, Z.; Gong, Y.; and Wei, X. 2024.
\newblock Artrackv2: Prompting autoregressive tracker where to look and how to describe.
\newblock In \emph{Proceedings of the IEEE/CVF Conference on Computer Vision and Pattern Recognition}, 19048--19057.

\bibitem[{Bertinetto et~al.(2016)Bertinetto, Valmadre, Henriques, Vedaldi, and Torr}]{SiamFC}
Bertinetto, L.; Valmadre, J.; Henriques, J.~F.; Vedaldi, A.; and Torr, P. H.~S. 2016.
\newblock Fully-Convolutional Siamese Networks for Object Tracking.
\newblock In \emph{{ECCV} Workshops}, 850--865.

\bibitem[{Bhat et~al.(2019)Bhat, Danelljan, Gool, and Timofte}]{DiMP50}
Bhat, G.; Danelljan, M.; Gool, L.~V.; and Timofte, R. 2019.
\newblock Learning Discriminative Model Prediction for Tracking.
\newblock In \emph{{ICCV}}, 6181--6190.

\bibitem[{Cai, Liu, and Wang(2024)}]{HIPTrack}
Cai, W.; Liu, Q.; and Wang, Y. 2024.
\newblock HIPTrack: Visual Tracking with Historical Prompts.
\newblock In \emph{Proceedings of the IEEE/CVF Conference on Computer Vision and Pattern Recognition}, 19258--19267.

\bibitem[{Chen et~al.(2023)Chen, Peng, Wang, Lu, and Hu}]{seqtrack}
Chen, X.; Peng, H.; Wang, D.; Lu, H.; and Hu, H. 2023.
\newblock SeqTrack: Sequence to Sequence Learning for Visual Object Tracking.
\newblock \emph{CVPR}, abs/2304.14394.

\bibitem[{Chen et~al.(2021)Chen, Yan, Zhu, Wang, Yang, and Lu}]{transt}
Chen, X.; Yan, B.; Zhu, J.; Wang, D.; Yang, X.; and Lu, H. 2021.
\newblock Transformer Tracking.
\newblock In \emph{{CVPR}}, 8126--8135.

\bibitem[{Chen et~al.(2020)Chen, Zhong, Li, Zhang, and Ji}]{Siamban}
Chen, Z.; Zhong, B.; Li, G.; Zhang, S.; and Ji, R. 2020.
\newblock Siamese Box Adaptive Network for Visual Tracking.
\newblock In \emph{{CVPR}}, 6667--6676.

\bibitem[{Cui et~al.(2022)Cui, Jiang, Wang, and Wu}]{mixformer}
Cui, Y.; Jiang, C.; Wang, L.; and Wu, G. 2022.
\newblock MixFormer: End-to-End Tracking with Iterative Mixed Attention.
\newblock In \emph{{CVPR}}, 13598--13608.

\bibitem[{Danelljan et~al.(2019)Danelljan, Bhat, Khan, and Felsberg}]{ATOM}
Danelljan, M.; Bhat, G.; Khan, F.~S.; and Felsberg, M. 2019.
\newblock {ATOM:} Accurate Tracking by Overlap Maximization.
\newblock In \emph{{CVPR}}, 4660--4669.

\bibitem[{Dosovitskiy et~al.(2021)Dosovitskiy, Beyer, Kolesnikov, Weissenborn, Zhai, Unterthiner, Dehghani, Minderer, Heigold, Gelly, Uszkoreit, and Houlsby}]{vit}
Dosovitskiy, A.; Beyer, L.; Kolesnikov, A.; Weissenborn, D.; Zhai, X.; Unterthiner, T.; Dehghani, M.; Minderer, M.; Heigold, G.; Gelly, S.; Uszkoreit, J.; and Houlsby, N. 2021.
\newblock An Image is Worth 16x16 Words: Transformers for Image Recognition at Scale.
\newblock In \emph{{ICLR}}.

\bibitem[{Fan et~al.(2021)Fan, Bai, Lin, Yang, Chu, Deng, Yu, Harshit, Huang, Liu, Xu, Liao, Yuan, and Ling}]{lasot-ext}
Fan, H.; Bai, H.; Lin, L.; Yang, F.; Chu, P.; Deng, G.; Yu, S.; Harshit; Huang, M.; Liu, J.; Xu, Y.; Liao, C.; Yuan, L.; and Ling, H. 2021.
\newblock LaSOT: {A} High-quality Large-scale Single Object Tracking Benchmark.
\newblock \emph{Int. J. Comput. Vis.}, 439--461.

\bibitem[{Fan et~al.(2019)Fan, Lin, Yang, Chu, Deng, Yu, Bai, Xu, Liao, and Ling}]{lasot}
Fan, H.; Lin, L.; Yang, F.; Chu, P.; Deng, G.; Yu, S.; Bai, H.; Xu, Y.; Liao, C.; and Ling, H. 2019.
\newblock LaSOT: {A} High-Quality Benchmark for Large-Scale Single Object Tracking.
\newblock In \emph{{CVPR}}, 5374--5383.

\bibitem[{Fu et~al.(2021)Fu, Liu, Fu, and Wang}]{STMTrack}
Fu, Z.; Liu, Q.; Fu, Z.; and Wang, Y. 2021.
\newblock STMTrack: Template-Free Visual Tracking With Space-Time Memory Networks.
\newblock In \emph{{CVPR}}, 13774--13783.

\bibitem[{Gao et~al.(2022)Gao, Zhou, Ma, Wang, and Yuan}]{aiatrack}
Gao, S.; Zhou, C.; Ma, C.; Wang, X.; and Yuan, J. 2022.
\newblock AiATrack: Attention in Attention for Transformer Visual Tracking.
\newblock In \emph{{ECCV} {(22)}}, 146--164.

\bibitem[{Gao, Zhou, and Zhang(2023)}]{GRM}
Gao, S.; Zhou, C.; and Zhang, J. 2023.
\newblock Generalized Relation Modeling for Transformer Tracking.
\newblock \emph{CVPR}, abs/2303.16580.

\bibitem[{Guo et~al.(2022)Guo, Zhang, Fan, Jing, Lyu, Li, and Hu}]{TransInMo}
Guo, M.; Zhang, Z.; Fan, H.; Jing, L.; Lyu, Y.; Li, B.; and Hu, W. 2022.
\newblock Learning Target-aware Representation for Visual Tracking via Informative Interactions.
\newblock In \emph{{IJCAI}}, 927--934.

\bibitem[{Huang, Zhao, and Huang(2021)}]{got10k}
Huang, L.; Zhao, X.; and Huang, K. 2021.
\newblock GOT-10k: {A} Large High-Diversity Benchmark for Generic Object Tracking in the Wild.
\newblock \emph{{IEEE} Trans. Pattern Anal. Mach. Intell.}, 43(5): 1562--1577.

\bibitem[{Kristan, Leonardis, and et.al(2020)}]{vot2020}
Kristan, M.; Leonardis, A.; and et.al. 2020.
\newblock The Eighth Visual Object Tracking {VOT2020} Challenge Results.
\newblock In \emph{{ECCV} Workshops {(5)}}, volume 12539 of \emph{Lecture Notes in Computer Science}, 547--601. Springer.

\bibitem[{Li et~al.(2019)Li, Wu, Wang, Zhang, Xing, and Yan}]{SiamRPN++}
Li, B.; Wu, W.; Wang, Q.; Zhang, F.; Xing, J.; and Yan, J. 2019.
\newblock SiamRPN++: Evolution of Siamese Visual Tracking With Very Deep Networks.
\newblock In \emph{{CVPR}}, 4282--4291.

\bibitem[{Lin et~al.(2017)Lin, Goyal, Girshick, He, and Doll{\'{a}}r}]{focalloss}
Lin, T.; Goyal, P.; Girshick, R.~B.; He, K.; and Doll{\'{a}}r, P. 2017.
\newblock Focal Loss for Dense Object Detection.
\newblock In \emph{{ICCV}}, 2999--3007.

\bibitem[{Lin et~al.(2014)Lin, Maire, Belongie, Hays, Perona, Ramanan, Doll{\'{a}}r, and Zitnick}]{coco}
Lin, T.; Maire, M.; Belongie, S.~J.; Hays, J.; Perona, P.; Ramanan, D.; Doll{\'{a}}r, P.; and Zitnick, C.~L. 2014.
\newblock Microsoft {COCO:} Common Objects in Context.
\newblock In \emph{{ECCV}}, 740--755.

\bibitem[{Mayer et~al.(2021)Mayer, Danelljan, Paudel, and Gool}]{keeptrack}
Mayer, C.; Danelljan, M.; Paudel, D.~P.; and Gool, L.~V. 2021.
\newblock Learning Target Candidate Association to Keep Track of What Not to Track.
\newblock In \emph{{ICCV}}, 13424--13434. {IEEE}.

\bibitem[{M{\"{u}}ller et~al.(2018)M{\"{u}}ller, Bibi, Giancola, Al{-}Subaihi, and Ghanem}]{trackingnet}
M{\"{u}}ller, M.; Bibi, A.; Giancola, S.; Al{-}Subaihi, S.; and Ghanem, B. 2018.
\newblock TrackingNet: {A} Large-Scale Dataset and Benchmark for Object Tracking in the Wild.
\newblock In \emph{{ECCV}}, 310--327.

\bibitem[{Rezatofighi et~al.(2019)Rezatofighi, Tsoi, Gwak, Sadeghian, Reid, and Savarese}]{giou}
Rezatofighi, H.; Tsoi, N.; Gwak, J.; Sadeghian, A.; Reid, I.~D.; and Savarese, S. 2019.
\newblock Generalized Intersection Over Union: {A} Metric and a Loss for Bounding Box Regression.
\newblock In \emph{{CVPR}}, 658--666.

\bibitem[{Vaswani et~al.(2017)Vaswani, Shazeer, Parmar, Uszkoreit, Jones, Gomez, Kaiser, and Polosukhin}]{attention}
Vaswani, A.; Shazeer, N.; Parmar, N.; Uszkoreit, J.; Jones, L.; Gomez, A.~N.; Kaiser, L.; and Polosukhin, I. 2017.
\newblock Attention is All you Need.
\newblock In \emph{{NIPS}}, 5998--6008.

\bibitem[{Voigtlaender et~al.(2020)Voigtlaender, Luiten, Torr, and Leibe}]{siamrcnn}
Voigtlaender, P.; Luiten, J.; Torr, P. H.~S.; and Leibe, B. 2020.
\newblock Siam {R-CNN:} Visual Tracking by Re-Detection.
\newblock In \emph{{CVPR}}, 6577--6587.

\bibitem[{Wang et~al.(2021)Wang, Zhou, Wang, and Li}]{trdimp}
Wang, N.; Zhou, W.; Wang, J.; and Li, H. 2021.
\newblock Transformer meets tracker: Exploiting temporal context for robust visual tracking.
\newblock In \emph{CVPR}, 1571--1580.

\bibitem[{Xie et~al.(2023)Xie, Chu, Li, Lu, and Ma}]{VideoTrack}
Xie, F.; Chu, L.; Li, J.; Lu, Y.; and Ma, C. 2023.
\newblock VideoTrack: Learning to Track Objects via Video Transformer.
\newblock In \emph{Proceedings of the IEEE/CVF Conference on Computer Vision and Pattern Recognition (CVPR)}, 22826--22835.

\bibitem[{Xie et~al.(2022)Xie, Wang, Wang, Cao, Yang, and Zeng}]{SBT}
Xie, F.; Wang, C.; Wang, G.; Cao, Y.; Yang, W.; and Zeng, W. 2022.
\newblock Correlation-Aware Deep Tracking.
\newblock In \emph{{CVPR}}, 8741--8750.

\bibitem[{Xie et~al.(2024)Xie, Zhong, Mo, Zhang, Shi, Song, and Ji}]{AQATrack}
Xie, J.; Zhong, B.; Mo, Z.; Zhang, S.; Shi, L.; Song, S.; and Ji, R. 2024.
\newblock Autoregressive Queries for Adaptive Tracking with Spatio-Temporal Transformers.
\newblock In \emph{Proceedings of the IEEE/CVF Conference on Computer Vision and Pattern Recognition}, 19300--19309.

\bibitem[{Xing et~al.(2023)Xing, Yifan, Yongchao, Dahu, and Yihong}]{ARTrack}
Xing, W.; Yifan, B.; Yongchao, Z.; Dahu, S.; and Yihong, G. 2023.
\newblock Autoregressive Visual Tracking.
\newblock In \emph{Proceedings of the IEEE/CVF Conference on Computer Vision and Pattern Recognition (CVPR)}, 9697--9706.

\bibitem[{Xu et~al.(2020)Xu, Wang, Li, Ye, and Yu}]{SiamFC++}
Xu, Y.; Wang, Z.; Li, Z.; Ye, Y.; and Yu, G. 2020.
\newblock SiamFC++: Towards Robust and Accurate Visual Tracking with Target Estimation Guidelines.
\newblock In \emph{{AAAI}}, 12549--12556.

\bibitem[{Yan et~al.(2021{\natexlab{a}})Yan, Peng, Fu, Wang, and Lu}]{stark}
Yan, B.; Peng, H.; Fu, J.; Wang, D.; and Lu, H. 2021{\natexlab{a}}.
\newblock Learning Spatio-Temporal Transformer for Visual Tracking.
\newblock In \emph{ICCV}, 10428--10437.

\bibitem[{Yan et~al.(2021{\natexlab{b}})Yan, Zhang, Wang, Lu, and Yang}]{Alpha-Refine}
Yan, B.; Zhang, X.; Wang, D.; Lu, H.; and Yang, X. 2021{\natexlab{b}}.
\newblock Alpha-Refine: Boosting Tracking Performance by Precise Bounding Box Estimation.
\newblock In \emph{{CVPR}}, 5289--5298. Computer Vision Foundation / {IEEE}.

\bibitem[{Ye et~al.(2022)Ye, Chang, Ma, Shan, and Chen}]{ostrack}
Ye, B.; Chang, H.; Ma, B.; Shan, S.; and Chen, X. 2022.
\newblock Joint Feature Learning and Relation Modeling for Tracking: {A} One-Stream Framework.
\newblock In \emph{{ECCV} {(22)}}, 341--357.

\bibitem[{Zhang et~al.(2019)Zhang, Gonzalez{-}Garcia, van~de Weijer, Danelljan, and Khan}]{updateNet}
Zhang, L.; Gonzalez{-}Garcia, A.; van~de Weijer, J.; Danelljan, M.; and Khan, F.~S. 2019.
\newblock Learning the Model Update for Siamese Trackers.
\newblock In \emph{{ICCV}}, 4009--4018.

\bibitem[{Zhang et~al.(2020)Zhang, Peng, Fu, Li, and Hu}]{Ocean}
Zhang, Z.; Peng, H.; Fu, J.; Li, B.; and Hu, W. 2020.
\newblock Ocean: Object-Aware Anchor-Free Tracking.
\newblock In \emph{{ECCV}}, 771--787.

\bibitem[{Zheng et~al.(2024)Zheng, Zhong, Liang, Mo, Zhang, and Li}]{odtrack}
Zheng, Y.; Zhong, B.; Liang, Q.; Mo, Z.; Zhang, S.; and Li, X. 2024.
\newblock Odtrack: Online dense temporal token learning for visual tracking.
\newblock In \emph{Proceedings of the AAAI Conference on Artificial Intelligence}, volume~38, 7588--7596.

\bibitem[{Zhou et~al.(2024)Zhou, Guo, Hong, Li, Zhang, Ge, and Zhang}]{RFGM}
Zhou, X.; Guo, P.; Hong, L.; Li, J.; Zhang, W.; Ge, W.; and Zhang, W. 2024.
\newblock Reading relevant feature from global representation memory for visual object tracking.
\newblock \emph{Advances in Neural Information Processing Systems}, 36.

\end{thebibliography}

\end{document}